\def\paperID{0000}
\def\confName{CVPR}
\def\confYear{2024}

\def\authorBlock{
    Lasse H. Hansen$^{1,2,}$\thanks{Equal contribution} \qquad
    Simon B. Jensen$^{1,3,}$\footnotemark[1] \qquad
    Mark P. Philipsen$^{1}$ \\
    Andreas Møgelmose$^{1}$ \qquad
    Lars Bodum$^1$ \qquad
    Thomas B. Moeslund$^{1}$ \\
    $^{1}$Aalborg University \qquad
    $^{2}$IT34 A/S \qquad
    $^{3}$Ambolt AI \\
    {\tt\small lhe@it34.com}, {\tt\small simon@ambolt.io}  {\tt\small \{mpph, anmo, tbm\}@create.aau.dk}, {\tt\small lbo@plan.aau.dk}\\
}

\newif\ifreview 
\newif\ifarxiv \newcommand{\arxiv}{\arxivtrue}
\newif\ifcamera 
\newif\ifrebuttal 

\arxiv  

\pdfoutput=1
\documentclass[10pt,twocolumn,letterpaper]{article}
\ifreview \usepackage[review]{cvpr} \fi
\ifarxiv \usepackage[pagenumbers]{cvpr} \fi
\ifrebuttal \usepackage[rebuttal]{cvpr} \fi
\ifcamera \usepackage{cvpr} \fi

\usepackage{graphicx}	
\usepackage{amsmath}	
\usepackage{amssymb}	
\usepackage{booktabs}
\usepackage{times}
\usepackage{microtype}
\usepackage{epsfig}
\usepackage[table,xcdraw,dvipsnames]{xcolor}
\usepackage{caption}
\usepackage{float}
\usepackage{placeins}
\usepackage{color, colortbl}
\usepackage{stfloats}
\usepackage{enumitem}
\usepackage{tabularx}
\usepackage{xstring}
\usepackage{multirow}
\usepackage{xspace}
\usepackage{url}
\usepackage{subcaption}
\usepackage{xcolor}
\usepackage[hang,flushmargin]{footmisc}

\usepackage{rotating} 
\usepackage{graphicx} 

\ifcamera \usepackage[accsupp]{axessibility} \fi

% ...

\ifarxiv  \fi

\newcommand{\R}[1]{{%
    \textbf{%
        \ifstrequal{#1}{1}{\textcolor{red}{R#1}}{%
        \ifstrequal{#1}{2}{\textcolor{blue}{R#1}}{%
        \ifstrequal{#1}{3}{\textcolor{magenta}{R#1}}{%
        \ifstrequal{#1}{4}{\textcolor{teal}{R#1}}{%
                           \textcolor{cyan}{R#1}%
        }}}}%
    }%
}}

\usepackage{xr-hyper}

\makeatletter
\newcommand*{\addFileDependency}[1]{
  \typeout{(#1)}
  \@addtofilelist{#1}
  \IfFileExists{#1}{}{\typeout{No file #1.}}
}

\makeatother
\newcommand*{\myexternaldocument}[1]{
    \externaldocument{#1}
    \addFileDependency{#1.tex}
    \addFileDependency{#1.aux}
}

\definecolor{cvprblue}{rgb}{0.21,0.49,0.74}
\usepackage[pagebackref,breaklinks,colorlinks,citecolor=cvprblue]{hyperref}
\usepackage[capitalize]{cleveref}
\crefname{section}{Sec.}{Secs.}
\crefname{table}{Table}{Tables}
\crefname{figure}{Fig.}{Figs.}

\ifarxiv \crefname{appendix}{App.}{Apps.}
\else \crefname{appendix}{Suppl.}{Suppls.} \fi

\unless\ifarxiv \myexternaldocument{_supplementary} \fi

\begin{document}
\title{OpenTrench3D: A Photogrammetric 3D Point Cloud Dataset for Semantic Segmentation of Underground Utilities}


\author{\authorBlock}
\maketitle

\begin{abstract}
Identifying and classifying underground utilities is an important task for efficient and effective urban planning and infrastructure maintenance.  
We present OpenTrench3D, a novel and comprehensive 3D Semantic Segmentation point cloud dataset, designed to advance research and development in underground utility surveying and mapping. 
OpenTrench3D covers a completely novel domain for public 3D point cloud datasets and is unique in its focus, scope, and cost-effective capturing method. 
The dataset consists of 310 point clouds collected across 7 distinct areas. These include 5 water utility areas and 2 district heating utility areas.
The inclusion of different geographical areas and main utilities (water and district heating utilities) makes OpenTrench3D particularly valuable for inter-domain transfer learning experiments.
We provide benchmark results for the dataset using three state-of-the-art semantic segmentation models, PointNeXt, PointVector and PointMetaBase. 
Benchmarks are conducted by training on data from water areas, fine-tuning on district heating area 1 and evaluating on district heating area 2.  
The dataset is publicly available 
\footnote{\href{https://github.com/SimonBuusJensen/OpenTrench3D}{https://github.com/SimonBuusJensen/OpenTrench3D}}.
With OpenTrench3D, we seek to foster innovation and progress in the field of 3D semantic segmentation in applications related to detection and documentation of underground utilities as well as in transfer learning methods in general. 
\end{abstract}

\begin{figure}[ht]
\centering
\includegraphics[width=1\linewidth]{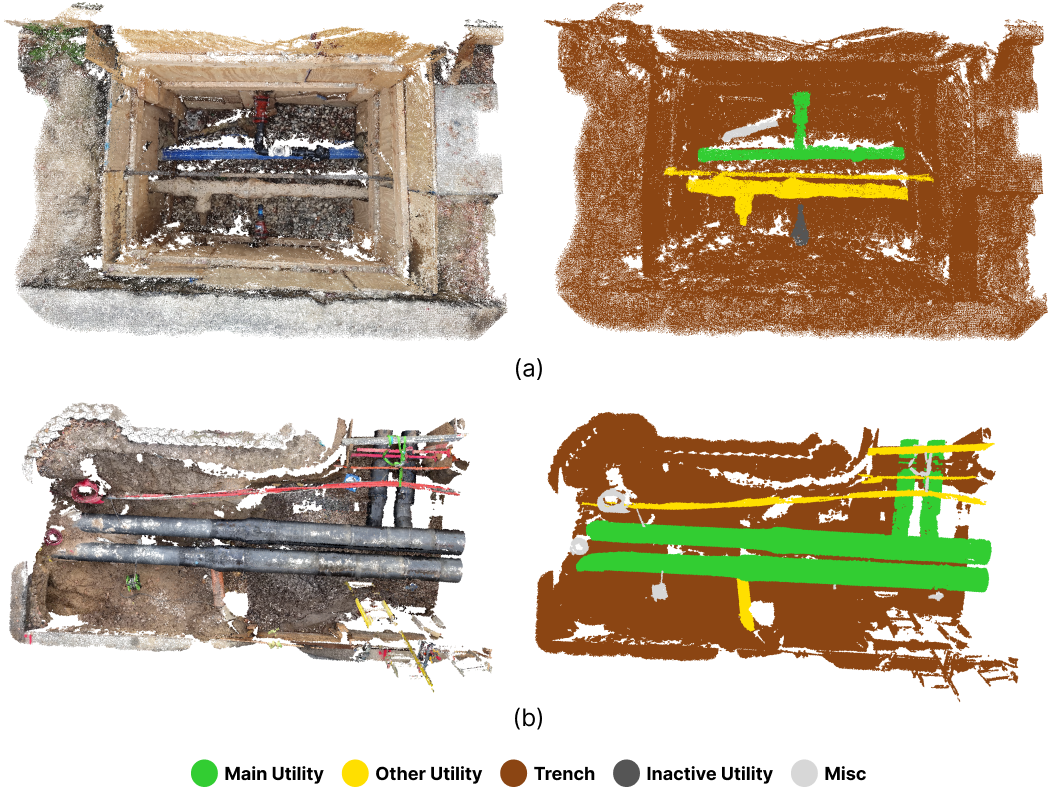}
\caption{Point clouds from OpenTrench3D: \textbf{(a)} In the water project areas the \textit{Main Utility} class is made up of water utility pipes. \textbf{(b)} In the district heating project areas the \textit{Main Utility} class is made up of district heating utility pipes. Find description of dataset and classes in section \ref{sec:dataset}.}
\label{fig:example-point-clouds}
\end{figure}

\section{Introduction}
\label{sec:intro}

Motivated by an ongoing and escalating demand for infrastructure development both above and beneath the ground, urban planners, engineers, and contractors depend on data regarding the placement of hidden underground utilities \cite{van_son_framework_2018, vinther2023visualisation}. 
Yet, a large portion of utility data is often inaccurate in location, outdated, and incomplete \cite{Hansen3Dgeoinfo2021-data, van_son_data_2019}. This lack of accurate and current information becomes particularly critical when considering the financial impact of excavation damage to underground infrastructure. 
Reports highlight that such damages cost more than GBP 200 million in the UK \cite{Strike2019} and USD 30 billion in the US \cite{dirt2019} alone in 2019. This underscores the necessity for utility owners to recognize the importance of consistently maintaining up-to-date and accurate utility map data.



Innovation in utility surveying and mapping therefore holds significant importance. The most common method is known as open trench surveying, which occurs after the installation of new utility lines or the replacement of existing ones, involving excavation on site. However, during these surveys, only the utilities of interest to the owner are documented, while other exposed utilities are frequently overlooked, as utility owners lack short-term incentives to re-survey, validate, and potentially report updates on the assets of other utility owners \cite{Hansen3Dgeoinfo2021-data}.

Following the recent trend in urban scene modeling, using affordable and accessible drone photogrammetry for 3D reconstruction of urban environments \cite{campus3d,sensat-urban,swiss3dcities,stpls3d}, the adoption of close-range photogrammetric data capture via off-the-shelf smartphones for utility surveying has gained popularity \cite{Hansen3Dgeoinfo2021-RC,yuen2022potential}. This is driven by the emergence of industry solutions like Pix4D. Unlike conventional open trench methods, such as direct single point measurement with GNSS receivers, this approach captures every piece of underground infrastructure visible in the 3D reconstruction. However, both the utilities of interest and other utilities must be digitized as polylines for subsequent integration into utility network systems such as GIS. The annotation process is manual and laborious. Developing methods for the classification and segmentation of utilities could not only greatly increase efficiency but also open new possibilities for streamlining the process of validating and updating the locations of existing utilities.

We introduce OpenTrench3D, to our knowledge, the first publicly available 3D point cloud dataset for the semantic segmentation of underground utilities captured in open trenches. The dataset was created from smartphone video recordings and reconstructed using photogrammetry, making it an accessible and affordable data acquisition method to replicate. It comprises 310 fully annotated point clouds, totaling approximately 528 million points across 5 label classes, following a utility owner-centric classification scheme. The dataset includes point clouds from water and district heating projects. One example of each is shown in Figure \ref{fig:example-point-clouds}.

Since OpenTrench3D mainly consists of point clouds from water projects, it offers an excellent opportunity to study transfer learning to the smaller set of district heating point clouds. We leverage water area point clouds to develop a pretrained model, which is then fine-tuned and tested on district heating areas. This method evaluates the potential of transfer learning for semantic segmentation in open trenches across diverse utility types. Hereby, we demonstrate to what extent the amount of newly annotated point clouds of the target main utility types affects performance. We summarize the contributions of this paper as follows:

\begin{enumerate}
    \item We present the first publicly available 3D point cloud dataset of underground utilities, OpenTrench3D, featuring 310 fully annotated point clouds and a utility owner-centric classification scheme.
    \item We benchmark state-of-the-art 3D semantic segmentation deep learning methods. The results highlight the potential and effectiveness of transfer learning across utility types and evaluate the generalizability of SOTA methods.
    \item We demonstrate that underground utility point clouds, acquired through cost-effective photogrammetry techniques, can be segmented using deep learning methods. Thus, making this method more accessible to providers and academia.
\end{enumerate}


\section{Related Work}
\label{sec:related}

\begin{table*}[ht]
\centering
\caption{A comparative overview of key urban datasets for 3D point cloud semantic segmentation. MLS: Mobile Laser Scanning system, TLS: Terrestrial Laser Scanning system, ALS: Aerial Laser Scanning system, UAV: Unmanned Aerial Vehicle and Ptgy: Photogrammetry.}
\label{table:comparison-datasets}
\begin{tabular}{@{}lcrrccccc@{}} 
\toprule\midrule
Dataset & Category & Capture Method  & Points & Area/Length & Classes & RGB \\
\midrule
S3DIS\cite{s3dis} & Indoor Scene-level & RGB-D & 273M & 6,020 m\textsuperscript{2} & 13 & Yes   \\
ScanNet\cite{scannet} & Indoor Scene-level & RGB-D  & 242M & 34,453 m\textsuperscript{2} & 20 & Yes  \\
Semantic3D\cite{semantic3d} & Street-level & TLS & 4,009M & 1 km & 9 & Yes   \\
Paris-rue-Madame\cite{paris-rue-madame} & Street-level  & MLS & 20M & 0.16 km & 17 & No  \\

SemanticKITTI\cite{semantickitti} & Street-level & MLS & 4,549M & 39.2 km & 28 & No   \\
Toronto-3D\cite{toronto3d} & Street-level  & MLS & 78M & 1 km & 9 & Yes  \\
DALES\cite{dales} & Aerial-level & ALS & 505M & 10 km\textsuperscript{2} & 9 & No  \\
Hessigheim 3D\cite{hessigheim3d} & Aerial-level & ALS & 126M & 0.19 km\textsuperscript{2}  & 11 & No   \\
SensatUrban\cite{sensat-urban} & Aerial-level & UAV Ptgy & 2,847M & 7.64 km\textsuperscript{2}  & 31 & Yes   \\
Swiss3DCities\cite{swiss3dcities} & Aerial-level & UAV Ptgy & 226M & 2.7 km\textsuperscript{2}  & 5 & Yes   \\
STPLS3D\cite{stpls3d}  & Aerial-level & UAV Ptgy & 150M & 1.27 km\textsuperscript{2}  & 6 & Yes \\\midrule
\textbf{OpenTrench3D (Ours)}  & \textbf{Underground Scene-Level}  & \textbf{Close-range Ptgy}  & \textbf{528M} & \textbf{3.814 m\textsuperscript{2}} & \textbf{5} & \textbf{Yes}\\

\midrule\bottomrule
\end{tabular}
\end{table*}

\subsection{Current 3D point cloud datasets}

Compared to existing 3D point cloud datasets, OpenTrench3D distinguishes itself as the first dataset specifically focused on open trench underground utilities. Despite its unique focus, we consider our dataset within the broader spectrum of 3D datasets for urban and outdoor environments. 
In this section, we offer a brief overview and comparison of existing point cloud datasets for semantic segmentation in this domain along with a comparative summary between OpenTrench3D and other well-known datasets in Table \ref{table:comparison-datasets}. Lastly, we briefly present current approaches for segmenting underground utilities.

Urban 3D datasets can generally be categorized based on their application area into two main groups: street-level and aerial-level. 

\textbf{Street-level:} Derived from Mobile Laser Scanning (MLS) and Terrestrial Laser Scanning (TLS), these datasets capture detailed urban features such as buildings, vehicles, and vegetation. MLS has been predominantly used in foundational datasets like SemanticKITTI\cite{semantickitti}, Paris-Lille-3D\cite{paris-lille-3d}, and Toronto-3D\cite{toronto3d}, whereas Semantic3D \cite{semantic3d} uses TLS. Recent additions to this category, such as Urban SGPCM \cite{urban-sqpcm} and SP3D \cite{saint-petersburg-3d}, maintain the core attributes of their predecessors while introducing variations in scale, class diversity and environmental context

\textbf{Aerial-level:} These datasets are obtained through Airborne Laser Scanning (ALS) and UAV-based photogrammetry, offering a bird's-eye view of urban structures on a large scale. Early examples such as LASDU\cite{lasdu} and DALES\cite{dales} use ALS, but UAV photogrammetry has since become the preferred method, as exemplified by Campus3D\cite{campus3d} and SensatUrban\cite{sensat-urban}. Similar to street-level datasets, newer aerial datasets primarily expand upon previous offerings by introducing variations in scale, class types, and specific urban environments (e.g., the HRHD-HK\cite{hrhd-hk} dataset focuses on high-rise buildings). 
 
While there is a wealth of 3D datasets for above-ground environments, there are no publicly available datasets for underground settings, despite the critical value and importance in urban planning \cite{van_son_framework_2018}. In the closest alternatives, we find 2D datasets like Sewer-ML \cite{haurum_sewer-ml_2021}, which contains annotations of defects in sewer inspection CCTV videos \cite{haurum_survey_2020}; however, this data, captured from inside underground pipes, is fundamentally different. OpenTrench3D addresses this gap by offering widely different 3D scenes of urban environments, specifically focusing on the infrastructure beneath street level. Moreover, what distinguishes OpenTrench3D from existing above-ground datasets is its method of data capture. Utilizing close-range photogrammetry with everyday smartphones sets it apart from the more commonly used laser scanning techniques and UAV photogrammetry. 
\subsection{Segmentation of underground utilities}

Semantic segmentation of 3D point clouds using point-based deep neural networks has gained traction following the pioneering work of PointNet and PointNet++ by Qi et al. \cite{qi2017pointnet, pointnet++}. Despite these advancements, no prior work has tackled the challenge of segmenting 3D point clouds of underground utilities with deep learning methods, likely due to the lack of available data.

A closely related and significant contribution is the work of Stranner et al. \cite{stranner_instant_2024}. They introduced a 3D fitting algorithm designed to align synthetically generated 3D cylinders with actual pipe structures, thereby updating the existing utility line maps to reflect the as-planned versus as-built states. Designed for on-site use, their system uses live 3D reconstructions from LiDAR-equipped mobile devices but requires user initiation via an Augmented Reality interface \cite{HansenISMAR2021}, leaving classification to the user. This approach, while innovative, underscores the need for more automated classification methods in 3D underground utility segmentation.


\section{OpenTrench3D Dataset}
\label{sec:dataset}


\subsection{Data Capture}

The point clouds in this dataset were captured using an end-to-end data capture and processing solution called 
SmartSurvey\footnote{\href{https://it34.com/en/services/smartsurvey-app-en/}{https://it34.com/en/services/smartsurvey-app-en/}}.
The solution uses close-range photogrammetry, more specifically Structure from Motion (SfM) with Multi-View Stereovision (MVS), employing video recordings from commercially available smartphones. The utility companies carried out the video capture process, while a surveying company assisted in geo-referencing the generated point clouds by carefully surveyed spray markings. For an illustrative example of the process, we refer to the supplemetary material. 
The point clouds are estimated to have an absolute geographic accuracy within ±5 cm \cite{Hansen3Dgeoinfo2021-RC}.



\subsection{Description of the dataset}
\label{subsec:description-dataset}

\begin{figure*}[ht]
    \centering
    \includegraphics[width=1\linewidth]{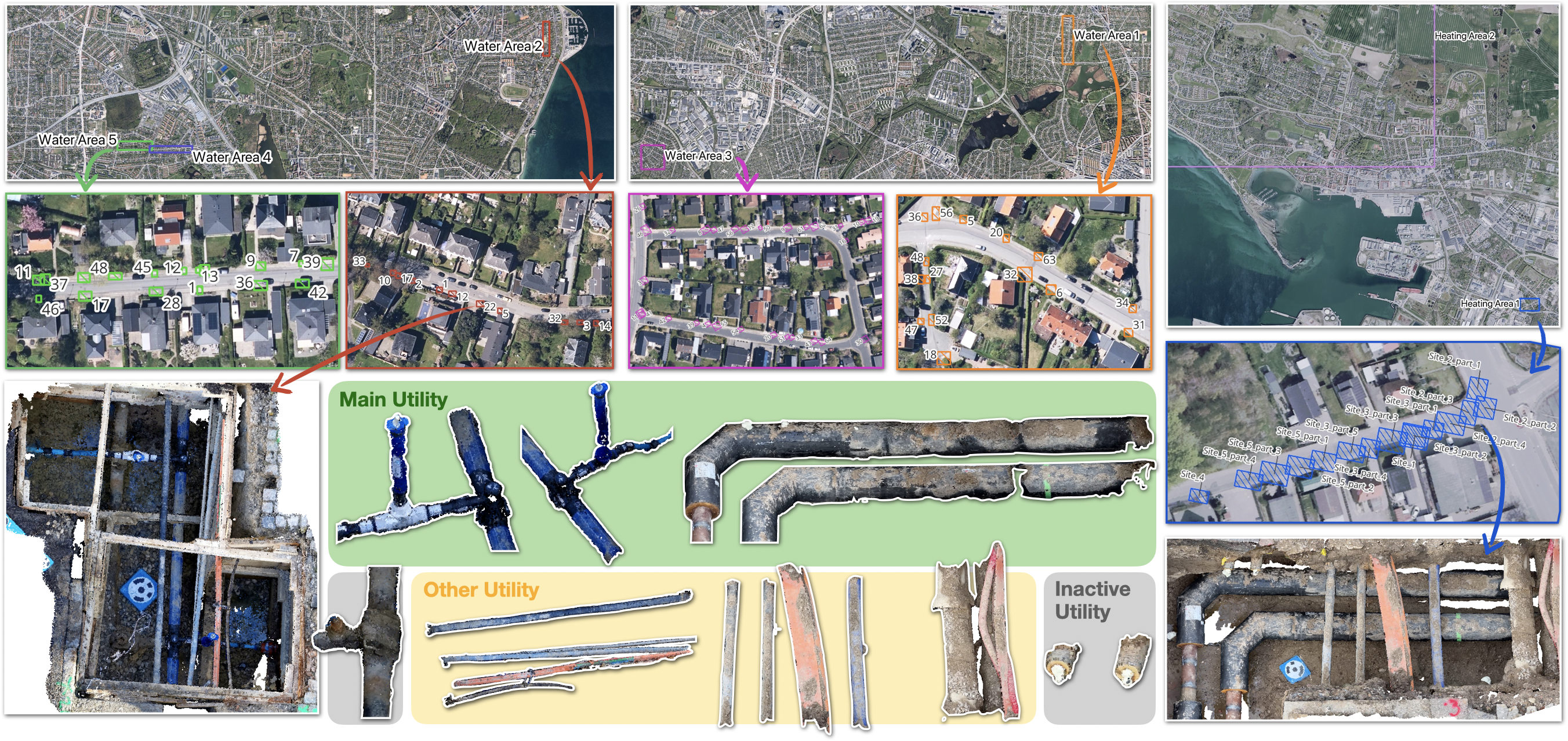}
    \caption{Map overview of the five water areas (top left and middle) and the two district heating areas (right), with zoom-in maps of selected parts of some areas. Two point cloud samples are presented from the water and heating areas, respectively, and utilities from the \textit{Main Utility}, \textit{Other Utility}, and \textit{Inactive Utility} classes are highlighted for comparison.}
    \label{fig:dataset-map}
\end{figure*}

The 3D point clouds, documenting newly installed pipes, comprise 310 point clouds from 7 areas. Five areas relate to water supply projects, while two are dedicated to district heating projects, each associated with different utility companies. These areas are named Water Area 1-5 and Heating Area 1-2, all located in different urban regions of Denmark as showed on the maps in Figure \ref{fig:dataset-map}.
The point clouds underwent subsequent post-processing, which included applying voxel downsampling to achieve a resolution of 4 mm, denoising using statistical outlier removal method, and manually removing noise artifacts in specific instances. Additionally, the background environment around the trench was removed to reduce the size of the point cloud. Using CloudCompare software \cite{CloudCompare2023}, the point clouds were manually segmented, with labels assigned to each point. The attributes for each point cloud PLY file are as follows: column 1-3: x, y and z-coordinates in meters. Column 4-6: r, g, b-color channels with values from 0-255. Colum 7: class id. 

\subsection{Utility Owner-centric classification scheme }

When deciding on a classification scheme for our dataset, we initially sought inspiration from existing standards and data models focusing on underground utility networks. Notable examples include the CityGML Utility Network ADE \cite{kutzner_semantic_2018} and UUDM \cite{yan_underground_2021}, which on a high-level provide classifications based on the utility type such as water, gas, electricity, sewage, telecommunications, etc. At first glance, adopting a similar classification scheme based on utility types appears logical. We discovered that accurately classifying utilities based on appearance is challenging, especially for older pipes, that are covered in dirt. This task is complex even for experts, who also gathers utility records from all relevant utility owners and carriers out on-site inspections. Gathering the extensive information required for confident classification exceeded the scope of this paper. 

Given these insights, we propose a utility owner-centric classification scheme that divides underground utilities into three separate classes, as showed in figure \ref{fig:dataset-map}, with additionally two classes for background and misc objects. These classes are naturally distinguishable from a utility owner's perspective and will be advantageous for subsequent utility mapping tasks. The five classes are defined as follows: 

\noindent \textbf{Trench:} The surrounding open excavation pit where the utilities are laid. The class include everything not described in the other classes. 

\noindent \textbf{Main Utility:} Newly installed utilities, which is the main utility of interest for surveying and mapping. In our dataset, this class is representing two distinct types of utilities: water and district heating. 
These utility types differ in both appearance and shape. 
For example, newly installed water pipes predominantly feature distinct blue colors, while district heating pipes are colored black. However, within the area of one utility owner, the newly installed utilities often consist of the same materials and components, giving them a recognisable look. 

\noindent \textbf{Other Utility:} Existing utilities uncovered during excavation belonging to other utility owners.
This class represents a very diverse set of utility structures but shares more similarities across all areas in the dataset.

\noindent \textbf{Inactive Utility:} Out-of-service utilities belonging to the utility owner, but are still left installed in the ground when replacing older counterparts of the utilities in the \textit{Main Utility} class. Due to the large time gap between installations, these older utilities vary in appearance—from closely resembling the newly installed utilities in the \textit{Main Utility} class to resembling structures in the \textit{Other Utility} class. Labelling this class proved challenging and was cross-referenced with utility map data provided by the same utility owners that also provided the point cloud data for the dataset.

\noindent \textbf{Misc:} Miscellaneous trench items such as pipe-like objects, work equipment and left-over cut pipe segments. These items are captured, due to the messy nature of an excavation site, but irrelevant for the utility owner to document.


\begin{table*}[ht]
\centering
\caption{The number of point clouds in each area and the number of points per class and the average number of points per point cloud. All numbers except point clouds are stated in thousands. The total number of points in the dataset is 528 million points.}
\label{table:dataset-stats}
\begin{tabular}{@{}lrlllllll@{}} 
\toprule\midrule

\multirow{2}{*}{Area} & \multirow{2}{*}{Clouds} &   Main & Other  & \multirow{2}{*}{Trench} & Inactive & Misc & Avg. \\
& & Utility& Utility & & Utility & & Points \\
\midrule
Water Area 1            & 68            &	2,566 (2.6\%) 	        &   3,377 (3.5\%) 	         & 90,578 (92.8\%)	        & 842 (0.9\%) & 195 (0.2\%)	                & 1,435  \\
Water Area 2            & 35            &	1,183 (3.2\%)	        &   1,540 (4.1\%) 	         & 33,735 (90.6\%)	         & 680 (1.8\%) & 80 (0.2\%)	                     & 1,063 \\
Water Area 3            & 65            &	3,380 (3.0\%)	        &   4,192 (3.8\%)	         & 101,970 (91.5\%)	         & 1,860 (1.7\%) & 64 (0.1\%)	               & 1,715 \\
Water Area 4            & 42            &	1,946 (3.0\%)	        &   3,816 (5.9\%)	         & 58,237 (90.4\%)	         & 358 (0.6\%) & 57 (0.1\%)	                     & 1,534 \\
Water Area 5            & 50            &	1,606 (2.2\%)	        &   3,403 (4.7\%)	         & 66,937 (92.5\%)	         & 377 (0.5\%) & 22 (0.0\%)	                     & 1,447 \\\midrule
\textbf{Water Total}    & \textbf{260}  &	\textbf{10,682 (2.8\%)}	&   \textbf{16,328 (4.3\%)}	 & \textbf{351,456 (91.8\%)}	 & \textbf{4,118 (1.1\%)} & \textbf{418 (0.1\%)} & \textbf{1,473} \\\midrule
Heat Area 1             & 29            &	20,363 (21.3\%)	        &   1,605 (1.7\%)	         & 72,927 (76.1\%)	         & 576 (0.6\%) & 321 (0.3\%)	                     & 3,303  \\
Heat Area 2             & 21            &	5,191 (10.5\%)	        &   535	(1.1\%)             & 43,252 (87.7\%)	         & 18 (0.0\%) & 326 (0.7\%)                     & 2,349  \\\midrule
\textbf{Heat Total}     & \textbf{50}   &	\textbf{25,554 (17.6\%)}	&   \textbf{2,140 (1.5\%)}	 & \textbf{116,179 (80.1\%)}	 & \textbf{594 (0.4\%)} & \textbf{646 (0.4\%)} & \textbf{2,902} \\\midrule
\bottomrule
\end{tabular}
\end{table*}

Table \ref{table:dataset-stats} shows the distribution of points for each class across all areas of the dataset. 
It shows a large overall class imbalance, with the \textit{Trench} class, being by far the most dominate class, as it serves as the background class. 
In water areas, the distribution among the three utility classes is relatively balanced, with the \textit{Main Utility} class consistently represented across all areas. 
A notable difference between the water and heating areas is the higher number of points in the \textit{Main Utility} class in heating areas, which aligns with the fact that district heating utilities are larger than water utilities. 

\section{Evaluation methods on OpenTrench3D}
\label{sec:evaluation}

\subsection{3D Semantic Segmentation Methods}
\label{subsec:semantic-segmentation-networks-for-benchmarking}
We provide an intial benchmark for OpenTrench3D using three state-of-the-art 3D semantic segmentation deep learning networks: PointNeXt\cite{pointnext}, PointVector\cite{pointvector} and PointMetaBase\cite{pointmetabase}. 
We choose three point-based methods which employs different feature extraction methodologies and achieve state-of-the-art performance on S3DIS\cite{s3dis} and ScanNetV2\cite{scannet}, which are scene-level datasets like OpenTrench3D:\\
\textbf{PointNeXt\cite{pointnext}}: modernises the highly influential PointNet++\cite{pointnet++} with improved training strategies e.g., data augmentation and optimization techniques and introduces the inverted residual bottleneck block which allows for efficient and effective model scaling. We use the XL version of PointNeXt which has 41.5M parameters.\\
\textbf{PointVector\cite{pointvector}}: enhances on the Point Set Abstraction blocks of the PointNet series, which are responsible for aggregating features of neighboring points. This is done by introducing a Vector-oriented Point Set Abstraction block that can aggregate neighboring features through higher-dimensional vectors instead of simple scalars. We use the XL version of PointVector which has 24.1M parameters\\
\textbf{PointMetaBase\cite{pointmetabase}}: introduces the concept of a PointMeta-building block, which is composed of a four sub-blocks: a neighbor update function, a neighbor aggregation function, a point update function and a position embedding function. PointMetaBase experiements with various settings for each sub-block to identify optimal feature-extraction capabilities while allowing for efficient processing of points. We use the XXL version of PointMetaBase which has 19.7M parameters.

\subsection{Evaluation Metrics}
We use two well-established metrics, mean intersection-over-union (mIoU) and mean accuracy (mAcc), for evaluating the 3D semantic segmentation performance of the methods. 
Intersection-over-union for class $i$ is given by:

\begin{equation}
IoU_i = \frac{\text{TP}_i}{\text{TP}_i + \text{FP}_i + \text{FN}_i}
\end{equation}

The mean intersection-over-union is given by:

\begin{equation}
mIoU = \frac{1}{N} \sum\nolimits_{i=0}^{N}{IoU_i}
\end{equation}

The accuracy for class $i$ is given by:

\begin{equation}
Acc_i = \frac{\text{TP}_i}{\text{TP}_i + \text{FN}_i}
\end{equation}

The mean accuracy is given by:

\begin{equation}
mAcc = \frac{1}{N} \sum\nolimits_{i=0}^{N}{Acc_i}
\end{equation}

\begin{figure*}[ht]
\centering
\includegraphics[width=1\linewidth]{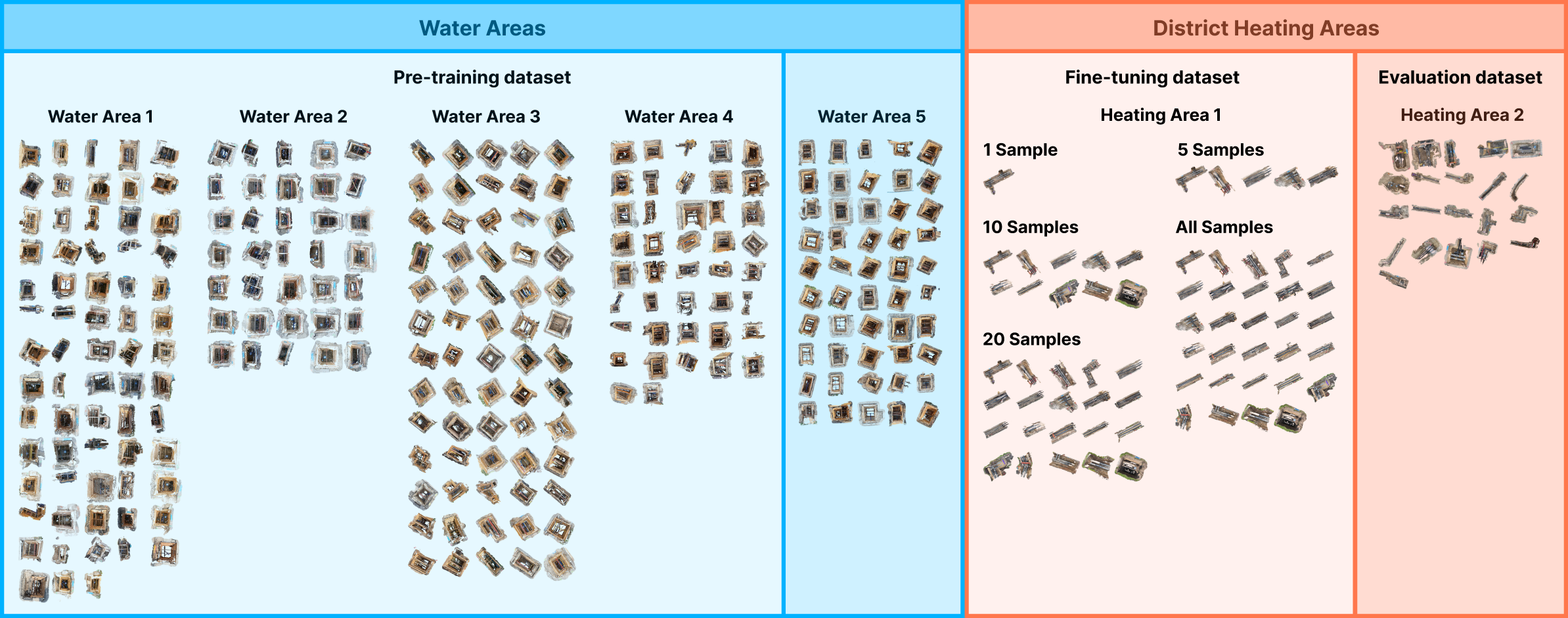}
\caption{An overview of the dataset sub-sets of OpenTrench3D used in the 5-fold cross-validation on water areas and fine-tuning evaluation on heating areas as described in section \ref{subsec:experiment-desc}.}
\label{fig:dataset-areas-overview}
\end{figure*}

\subsection{OpenTrench3D Evaluation Description}
\label{subsec:experiment-desc}

\noindent \textbf{5-fold cross-validation on water areas:} We conduct a 5-fold cross-validation on the water areas in two rounds. Initially, we include the \textit{Inactive Utility} class during training and evaluation and subsequently, we ignore it. This allows us to assess the three methods' effectiveness in distinguishing among the three utility classes. 
Additionally, the \textit{Misc} class is ignored from both evaluations due to its lack of relevance to the Utility Owner.

\noindent \textbf{Fine-tuning evaluation on heating areas:} We conduct a fine-tuning evaluation to asses the transfer learning and fine-tuning opportunities of the dataset and to investigate the generalizability and performance of the three methods on the entire dataset.
Specifically, we first pre-train model weights by training on Water Area 1-4 while using Water Area 5 as a validation set for the pre-training.
Secondly, we fine-tune the model weights on 1, 5, 10, 20 and all (29) samples from Heating Area 1. 
We conduct fine-tuning experiments in which only weights of the segmentation head of each model are fine-tuned as well as experiments in which the weights of both the segmentation head and the decoder are fine-tuned, simultaneously. 
Finally, the fine-tuned models are evaluated on point clouds from Heating Area 2.  

An overview of the dataset subsets is presented in figure \ref{fig:dataset-areas-overview}.
In the fine-tuning evaluation, we ignore the \textit{Inactive Utility} class due to its minimal presence in heating area 2. 
The \textit{Misc} class is ignored in all experiments.


\subsection{Training Environment}
The experiments for the 5-fold evaluation on water areas and fine-tuning evaluation on heating areas are run on a high-performance computing cluster with Nvidia A40 GPUs (48GB). 
In each experiment, the deep learning networks are assigned 64GB RAM, 8 CPU Cores (AMD EPYC) and 1 Nvidia A40 GPU. For the development environment, we use Ubuntu 20.04, python 3.10 and torch 1.10 with CUDA 11.3. 

Pre-training experiments are run for 50 epochs with a batch size of 8. For each epoch of training we loop over the dataset 30 times similar to \cite{pointnext}. 
The voxel max, which specifies the number of processed points per point cloud per batch, is set to 24,000. 
We use a Adamw optimizer with an initial learning rate of 0.01 and a minimum learning rate of 0.0001. 
Fine-tuning experiments are run for 50 epochs with a batch size of 1 when fine-tuning on 1 and 5 samples, a batch of 4 when fine-tuning on 10 samples and a batch size of 8 when fine-tuning on 20 and 29 samples. The learning rate is set to 0.05 and minimum learning rate of 0.0001. 
We universally down-sample point clouds using a voxel grid size of 0.02m and set the query ball radius to 0.025m. 
These settings are derived from using data-driven hyperparameter-tuning \cite{data-driven-hyperparameter-tuning} and an additional hyperparameter sweep.

\begin{figure*}[ht]
\centering
\includegraphics[width=1\linewidth]{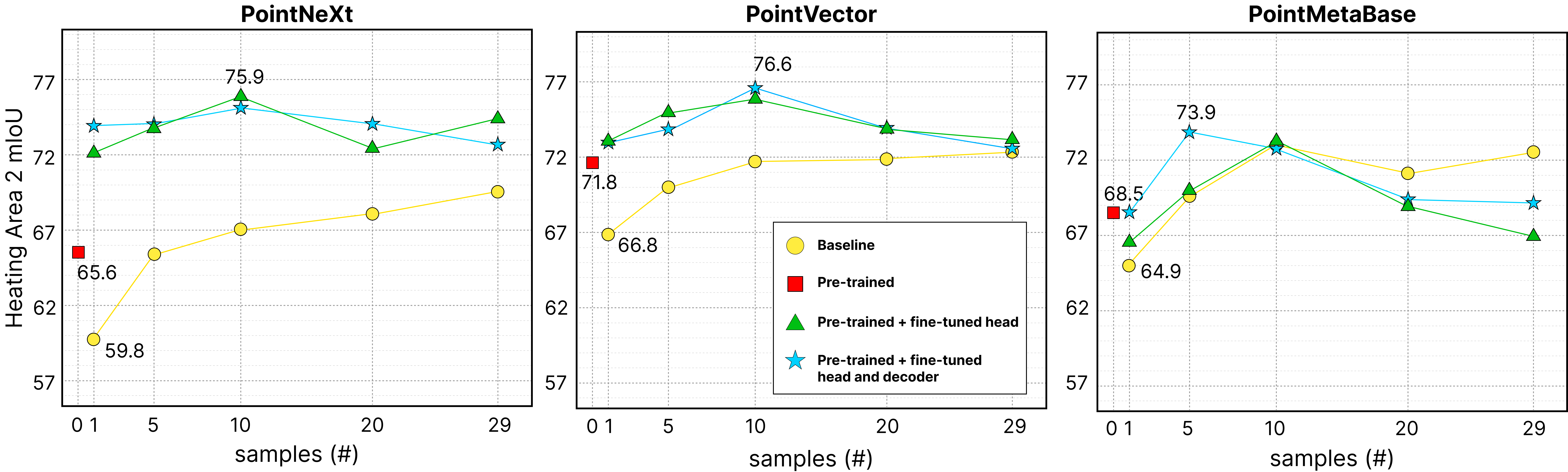}
\caption{Results from the fine-tuning experiments described in section \ref{subsec:description-dataset}. For comparison, we display results from pre-trained models (red square) and results of models solely trained on samples from Heating Area 1 (Baseline).}
\label{fig:results-fine-tuning-experiments}
\end{figure*}

\begin{table}[ht]
    \centering
    \caption{Results from 5-fold cross-validation on water areas as described in \ref{subsec:experiment-desc}. The mACC, mIoU and IoU score for each class.}
    \label{tab:5-fold-water}
    \begin{tabular}{@{}lcccccc@{}}
        \toprule
        \midrule
        Method          & \rotatebox[origin=c]{90}{mAcc} & \rotatebox[origin=c]{90}{mIoU} & \rotatebox[origin=c]{90}{\textit{Main}} & \rotatebox[origin=c]{90}{\textit{Other}} & \rotatebox[origin=c]{90}{\textit{Trench}} & \rotatebox[origin=c]{90}{\textit{Inactive}} \\ 
        \midrule
        PointNeXt       & 79.7 & 70.6 & 81.6               & 59.7                & 98.1         & 42.8   \\                
        PointVector     & 84.1 & \textbf{76.5} & 83.1               & \textbf{77.7 }               & \textit{98.6 }        & \textbf{46.4}   \\                
        PointMetabase   & \textbf{84.5} & 75.8 & \textbf{83.6}               & 76.6                & \textit{98.6}         & 44.5  \\ \midrule                 
        PointNeXt       & 91.9 & 88.5 & 87.8               & 79.1                & 98.7         & -  \\                    
        PointVector     & 93.2 & 90.3 & 90.3               & 81.8                & \textbf{98.8}         & -     \\                 
        PointMetabase   & \textbf{93.8} & \textbf{90.5} & \textbf{90.4}               & \textbf{82.2}                & \textbf{98.8}         & -     \\                  
        \midrule
        \bottomrule
    \end{tabular}
\end{table}

\begin{figure*}[ht]
\centering
\includegraphics[width=1\linewidth]{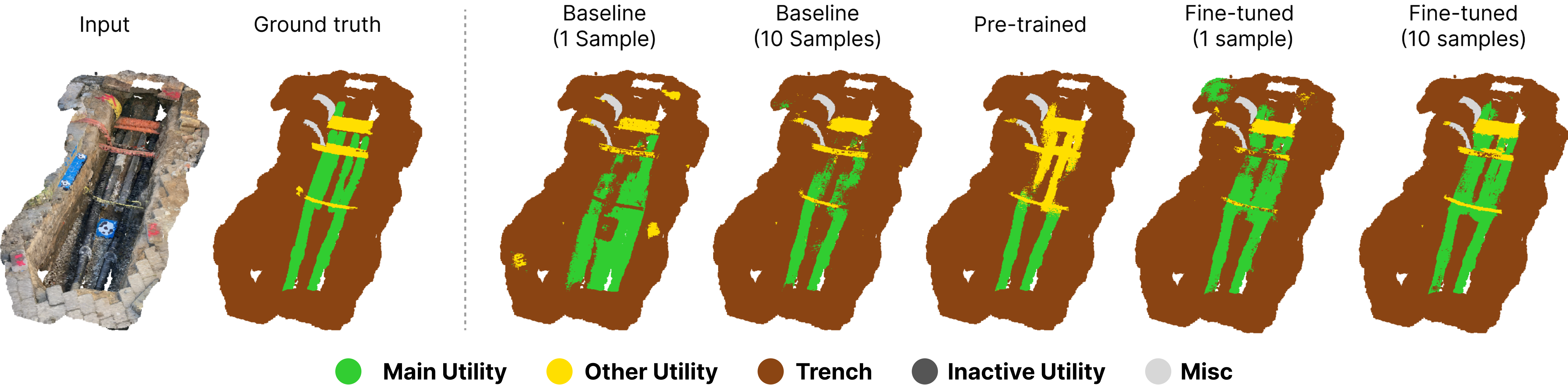}
\caption{Qualitative results of pre-trained PointNeXt model, PointNeXt models trained on 1 and 10 samples from Heat Area 1 and finally pre-trained and fine-tuned PointNeXt models fine-tuned on 1 and 10 samples were only weights of the segmentation head are tuned. A set of additional examples are seen in the supplementary material.}
\label{fig:fine-tuning-vs-training-qualitative}
\end{figure*}


\begin{table}[ht]
    \centering
     \caption{Results from fine-tuning evaluation on heating areas as described in \ref{subsec:experiment-desc} for PointNeXt. The mACC, mIoU and IoU score for each class is shown. We refer to the supplementary material for corresponding tables for PointVector and PointMetaBase.}
    \label{tab:pretraining-vs-baseline-vs-finetuning}
    \begin{tabular}{@{}lcccccc@{}}
        \toprule
        \midrule
        \textbf{PointNeXt} & \rotatebox[origin=c]{90}{Samples} & \rotatebox[origin=c]{90}{mAcc} & \rotatebox[origin=c]{90}{mIoU} & \rotatebox[origin=c]{90}{\textit{Main}} & \rotatebox[origin=c]{90}{\textit{Other}} & \rotatebox[origin=c]{90}{\textit{Trench}} \\
        \midrule
        \multirow{5}{*}{Baseline}       & 1     & 68.3  & 59.8  &  65.1 & 20.4  & 93.8  \\       
                                        & 5     & 72.8	& 65.6	& 78.7	& 21.3	& 96.9  \\
                                        & 10    & 72.7	& 67.1	& 75.8	& 29.2	& 96.3  \\ 
                                        & 20    & 76.2	& 68.1	& 75.9	& 32.1	& 96.4  \\ 
                                        & 29    & 75.5	& 69.6	& 79.5	& 32.2	& 96.9  \\  
        \midrule
        Pre-trained &           & 75.0	& 65.6	& 70.6	& 29.2	& 97.0  \\
        \midrule
        \multirow{5}{*}{\parbox{2cm}{Fine-tuned\\(Head)}}     & 1   & 80.4	& 72.2	& 76.4	& 43.9	& 96.5        \\       
                                        & 5   & 79.8	& 73.8	& 81.2	& 43.0	& 97.2        \\
                                        & 10  & 83.7	& \textbf{75.9}	& 83.6	& \textbf{46.5}	& \textbf{97.6}        \\ 
                                        & 20  & 79.9	& 72.3	& 83.0	& 36.4	& \textbf{97.6}        \\ 
                                        & 29  & \textbf{84.5}	& 74.4	& 81.4	& 44.2	& 97.5        \\\midrule
        \multirow{5}{*}{\parbox{2cm}{Fine-tuned\\(Decoder+Head)}} &  1   &  80.8	& 73.9 &	80.2 &	 44.4 &	97.2
        \\       
                                        & 5   &  78.9 &	74.0 &	81.3 &	43.4 &	97.2
                                        \\
                                        & 10  &  82.0 &	75.1 &	\textbf{83.8} &	43.9 &	\textbf{97.6}
                                        \\ 
                                        & 20  &  83.9 &	74.1 &	82.4 &	42.4 &	97.5
                                        \\ 
                                        & 29  &  82.3 &	72.7 &	80.0 &	40.6 &	97.5
                                        \\ 
        \midrule
        \bottomrule
    \end{tabular}
\end{table}


\section{Results and Discussion}
\label{sec:discussion}

\subsection{5-fold cross-validation on Water Areas}
\label{subsec:5-fold-eval-water-results}
We present the outcomes of our 5-fold cross-validation in table \ref{tab:5-fold-water}. 
The \textit{Inactive Utility} class presents the most significant challenge, as expected, likely due to its visual similarities to \textit{Main Utility} and \textit{Other Utility} as visualized in zoomed in class examples of figure \ref{fig:dataset-map}. 
In our qualitative analysis, we note that the Inactive class is often misclassified as either Main or Other Utility. For illustrative examples of this, we direct readers to the supplementary materials.

\textit{Other Utility} emerges as the second most challenging class, whereas \textit{Main Utility} is the easiest to classify. 
Remarkably, in the second evaluation where \textit{Inactive Utility} is ignored, the IoU score of \textit{Main Utility} exceeds 90\% for both PointVector and PointMetabase, a level of performance that is difficult to enhance further and inspire the possibilities to build application upon for automating utility mapping. 
Our qualitative findings reveal that errors predominantly occur in boundary areas between two classes. In our dataset, this particularly happens in areas where one of the utility classes merge into the soil e.g., the \textit{Trench} class or when soil partially covers the utilities.

Generally, we notice a better performance in PointVector and PointMetabase compared to PointNeXt. This follows a similar trend in benchmarks on S3DIS dataset\cite{pointmetabase, pointvector}. 


\subsection{Fine-tuning evaluation on Heating Areas}
\label{subsec:results}
We present the results from the fine-tuning evaluation on heating areas in figure \ref{fig:results-fine-tuning-experiments} and table \ref{tab:pretraining-vs-baseline-vs-finetuning}. 
Generally, we observe that fine-tuning of pre-trained weights proves effective across all three methods evaluated, particularly when fine-tuning with 1, 5 and 10 samples from Heating Area 1.

We find PointVector performs the best, achieving a mIoU score of 76.6\% when fine-tuning the weights of both the segmentation head and the decoder with 10 samples from Heating Area 1, closely followed by PointNeXt with a mIoU score of 75.9\%.
Furthermore, PointVector consistently achieves the highest mIoU scores when pre-trained only on water areas. This supports the authors arguments that the vector-oriented Point Set Abstraction design introduces less inductive bias, which in returns improves generalization capabilities \cite{pointvector}.
One noteable results, is the fact that for both PointNeXt and PointVector, fine-tuning pre-trained weights with just a single sample from Heating Area 1 achieves a better performance compared to baseline version trained on even all 29 samples from Heating Area 1.
In contrast, PointMetaBase ranks as the least effective, with its baseline model trained on 20 and 29 samples even surpassing the performance of its fine-tuned counterparts.

Unexpectedly, we observe a negative trend upon increasing the number of fine-tuning samples to 20 and 29 for all three methods. 
This prompts the hypothesis that further experimental investigations into the optimization of hyper-parameter settings for fine-tuning with 20 and 29 samples to effectively capitalize on the additional data.


Compared to the 5-fold cross-validation on water areas, the mIoU score is significantly lower across all types of experiments, suggesting that achieving such performance likely requires more annotated data. 
However, this trend is less pronounced for the IoU score of the \textit{Main Utility} class, which still manages to achieve impressively high performance in the fine-tuning evaluation.
PointVector emerges as the top performer with an IoU score of 86.7\%, closely followed by PointNeXt with an IoU of 83.8\%, as seen in Table \ref{tab:pretraining-vs-baseline-vs-finetuning}. 
Both scores are relatively close to their counterparts in table \ref{tab:5-fold-water}. Figure \ref{fig:fine-tuning-vs-training-qualitative} clearly illustrates the significant improvement that fine-tuning brings to the \textit{Main Utility }class. When comparing models pre-trained solely on water areas to those fine-tuned on just 1 and 10 samples, a clear improvement is evident. The fine-tuning process drastically reduces the instances of the \textit{Main Utility} being incorrectly labeled as \textit{Other Utility}, achieving near-perfect results for both classes. The incorrect classification of the \textit{Main Utility} as \textit{Other} is further observed in our qualitative analysis and was expected, considering the visual and dimensional differences between water utilities and district heating pipes, as illustrated in Figure \ref{fig:dataset-map}.

Moreover, our qualitative analysis reveals surprising classification errors, in which the \textit{Trench} class is incorrectly identified as the Main Utility class, particularly in our fine-tuned and baseline models, and significantly less so in the model pre-trained on water areas. While the exact cause is uncertain, we hypothesize that the large difference in number of points, percentage-wise, in Heating Area 1 compared to other areas, may be a factor. Additionally, we observed that many of the heating pipes in Heating Area 1 are partially, and sometimes almost completely, covered with soil, which might contribute to the confusion. 

Upon examining the incorrect predictions, the error often appear obvious to the human eye. 
For instance, where a single continuous pipe is classified as both \textit{Main-} and \textit{Other Utiltiy} classes at different segments of the pipe. 
We believe this calls for post-processing, possibly integrated with some prior knowledge about class appearances or human interaction related to the work by Stranner et al. \cite{stranner_instant_2024}.



Finally, from the fine-tuning evaluation experiment, it has been discerned that the batch size has a significant influence on the methods' performance, particularly when fine-tuning with few samples e.g. 5 and 10. 
For example, upon fine-tuning a pre-trained PointVector model on 10 samples with a batch size of 1 over 50 epochs, a mIoU of 66.6\% was attained, whereas under identical conditions, when adjustment to a batch size of 4, the mIoU score was 76.6\%. A difference of 10.0 percentage points. 
Comparable trends were observed for PointMetaBase and PointNeXt.

\section{Conclusion}
\label{sec:conclusion}

In this paper, we introduce OpenTrench3D, the first publicly available point cloud dataset of underground utilities from open trenches.
It features 310 fully annotated point clouds consisting of a total of 528 million points categorised into 5 unique classes following a utility owner-centric classification scheme.
The dataset is acquired through photogrammetric techniques, offering an accessible and cost-effective method for capturing underground utility data.
Our experiments on OpenTrench3D involve: (i) a 5-fold cross-validation on water project areas, highlighting impressive performance for the main utility classes using state-of-the-art deep learning semantic segmentation methods and (ii) fine-tuning pre-trained models on district heating utilities with minimal samples, demonstrating the capability to learn segmenting entirely new utility classes effectively from just a few examples.
By providing this dataset, we aim to catalyze innovation and advance the research in the domain, ultimately contributing to reducing excavation damages, optimizing urban planning and improving underground utility surveying and mapping.

\section{Acknowledgement}
\label{sec:acknowledgement}
We thank the students who assisted in data annotation, and to IT34, Ambolt AI, Innovation Fund Denmark, DigitalLead and AI Denmark for their support and funding. Special thanks to Kalundborg and Novafos Utility Company for granting data access. 
\newpage

{\small
\bibliographystyle{ieeenat_fullname}
\bibliography{11_references}
}

\ifarxiv \clearpage \appendix \section{Introdution}
This document presents the supplementary materials
omitted from the main paper due to the space limitation.
 
\section{Data Capture Process}
\label{appendix_a}

The OpenTrench3D dataset is gathered using close-range photogrammetry captured using video recordings from everyday smartphones. The following is a description of the overall data capture and processing workflow used by the two utility owners that provided point clouds for OpenTrench3D. We refer to figure \ref{fig:data-capture_v2} for illustrations. This workflow highlights that utility owners on site only requires a marker and a smartphone to fulfill their role in the data capture process. The procedure is divided into three straightforward steps: (1) apply markings around the open trench, used as Ground Control Points (GCP), possibly using a spray marker; (2) carefully video record the trench from various angles, ensuring the camera is aimed down towards the utilities visible in the trench; and (3) upload the captured video through the companion application. 
Subsequent, the video data is then send to a server for processing into a 3D point cloud. Following the initial step, a surveying responsible is tasked with measuring the GCP markings using survey-grade instruments, such as GNSS-RTK receivers followed by uploading this data to the same system for later manual geo-referencing of the point cloud. 

\begin{figure}[ht]
    \centering
    \includegraphics[width=1\linewidth]{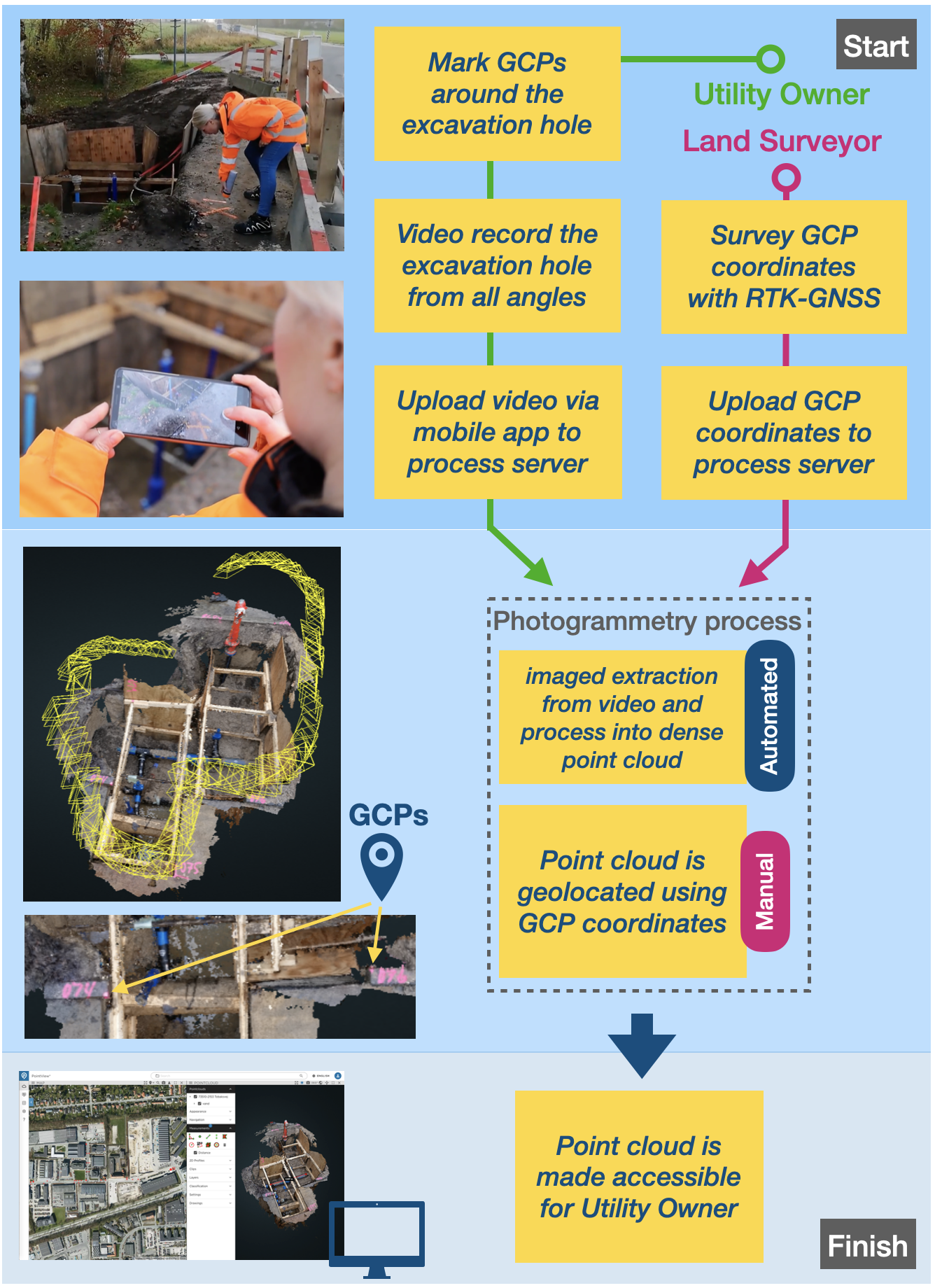}
    \caption{Workflow of the data capture process to generate 3D point cloud using photogrammetry. GCP stands for Ground Control Point.}
    \label{fig:data-capture_v2}
\end{figure}

\section{Fine-tuning Evaluation on Heating Areas}
\label{sec:appendix_b}
In table \ref{tab:supplementary-pretraining-vs-baseline-vs-finetuning} we present the supplementary results of the fine-tuning evaluation on heating areas of PointVector and PointMetaBase, similar to table 4 in the main article for PointNeXt. 
In the fine-tuning evaluation on heating areas, we first pre-train model weights on samples from Water Area 1-4 while using Water Area 5 as validation set for the pre-training. 
Secondly, we fine-tune the model weights on 1, 5, 10, 20 and all (29) samples from
Heating Area 1. 
We both conduct fine-tuning experiments in which only weights of the segmentation head of each model are fine-tuned as well as experiments in which the weights of both the segmentation head and the decoder are fine-tuned, simultaneously. 
Finally, the fine-tuned models are evaluated on point clouds from Heating Area 2.

\begin{table*}[ht]
    \centering
    \caption{Supplementary table for the fine-tuning experiments seen in figure 4 of the main paper. For comparison, we display the performance of pre-trained models (red square) as well as models without prior pre-training, but solely trained on 1, 5, 10, 20 and all samples from Heating Area 1 (Baseline).}
    \label{tab:supplementary-pretraining-vs-baseline-vs-finetuning}
    \begin{tabular}{@{}lc|ccccc|ccccc@{}}
        \toprule
        \midrule
        & & \multicolumn{5}{|c|}{\textbf{PointVector}} & \multicolumn{5}{|c}{\textbf{PointMetaBase}}  \\ 
        \midrule
         & \rotatebox[origin=c]{90}{Samples} & \rotatebox[origin=c]{90}{mAcc} & \rotatebox[origin=c]{90}{mIoU} & \rotatebox[origin=c]{90}{Main} & \rotatebox[origin=c]{90}{Other} & \rotatebox[origin=c]{90}{Trench} & \rotatebox[origin=c]{90}{mAcc} & \rotatebox[origin=c]{90}{mIoU} & \rotatebox[origin=c]{90}{Main} & \rotatebox[origin=c]{90}{Other} & \rotatebox[origin=c]{90}{Trench}\\
        \midrule
        \multirow{5}{*}{Baseline}   & 1     &  70.9	&   66.8	& 78.1	&  25.5	 & 96.7    &   70.3 &	64.9&	73.2 &	25.6 &	96.0    \\       
                                    & 5     &  73.8	&   70.0	& 77.4	&  36.0	 & 96.6    &   74.6 &	69.4&	75.4 &	36.4 &	96.5    \\
                                    & 10    &  76.2	&   71.7	& 79.8	&  38.4	 & 97.0    &   77.1 &	73.0&	78.8 &	\textbf{43.2} &	97.0    \\ 
                                    & 20    &  79.5	&   71.8	& 79.8	&  38.6	 & 97.1    &   \textbf{80.4} &	71.2&	79.9 &	36.3 &	97.5    \\ 
                                    & 29    &  81.2	&   72.6	& 77.5	&  43.1	 & 97.1    &   79.5 &	72.5&	80.3 &	39.8 &	97.2    \\  
        \midrule
        Pre-trained &               &  79.3 &  71.8 &   80.3	& 37.6	& 97.6   & 76.7 &	68.5 &	75.7 &	32.4 &	97.4      \\
        \midrule
        \multirow{5}{*}{\parbox{2cm}{Fine-tuned\\(Head)}}            & 1     &  76.4	&   73.1	& 83.8	&  37.9	 & 97.7   & 72.6 &	66.6&	76.0 &	27.5 &	96.3  \\       
                    & 5     &  77.4	&   74.0	& 82.8	&  41.8	 & 97.5    & 74.3 &	70.0&	75.4 &	37.9 &	96.6   \\
                    & 10    &  82.5	&   75.9	& 85.8	&  44.0	 & 97.9    & 78.2 &	73.3 &	81.4	 & 41.0 &	97.4 \\ 
                    & 20    &  81.8	&   73.8	& 83.5	&  39.9	 & 98.0    & 79.7 &	68.9&	78.4 &	30.7 &	\textbf{97.7}  \\ 
                    & 29    &  \textbf{84.5}	&   73.2	& 84.6	&  37.3	 & 97.8      & 79.5 &	66.9&	72.0 &	31.4 &	97.3 \\ 
        
        \midrule
        \multirow{5}{*}{\parbox{2cm}{Fine-tuned\\(Decoder+Head)}} & 1     &   76.2	&   73.0	& 83.8	&  37.7	 & 97.6     &  75.9 &	68.6&	72.7 &	37.4 &	95.8    \\   
                    & 5     &   77.5	&   73.9	& 83.3	&  40.7	 & 97.6     & 78.8 &	\textbf{73.9} &	\textbf{81.8} &	42.6 &	97.4     \\
                    & 10    &   82.2	&   \textbf{76.6}	& \textbf{86.7}	&  \textbf{45.2}	 & \textbf{98.1}     & 79.0 &	72.8&	80.9 &	40.0 &	97.4   \\ 
                    & 20    &   81.8	&   74.0	& 84.3	&  39.5	 & \textbf{98.1}      & 80.0 &	69.4&	78.9 &	31.4 &	\textbf{97.7}    \\ 
                    & 29    &   80.1	&   72.6	& 81.3	&  38.9	 & 97.8      &  78.9 &	69.1&	75.4 &	34.7 &	97.3   \\ 
        
        \midrule
        \bottomrule
    \end{tabular}
\end{table*}

\section{Qualitative Examples from 5-fold cross-validation on Water Areas}
\label{sec:appendix_c}
In figure \ref{fig:5-fold-inactive-large}, we provide qualitative results from running inference on various samples from Water Area 5 using a PointVector model trained on Water Area 1-4 with the \textit{Inactive Utility} class included (colored in blue) for evaluation.
The qualitative examples highlight the challenges state-of-the-art semantic segmentation methods encounter when trying to distinct the \textit{Inactive Utility} against the \textit{Main Utility} and \textit{Other Utility} classes.
Although, sometimes succesful, often times the methods neglect the \textit{Inactive Utility} class, possible due to it occuring less frequently in the dataset compared to the other classes, as seen from table 2 in the main paper. 

\begin{figure*}[ht]
\centering
\includegraphics[width=0.85\linewidth]{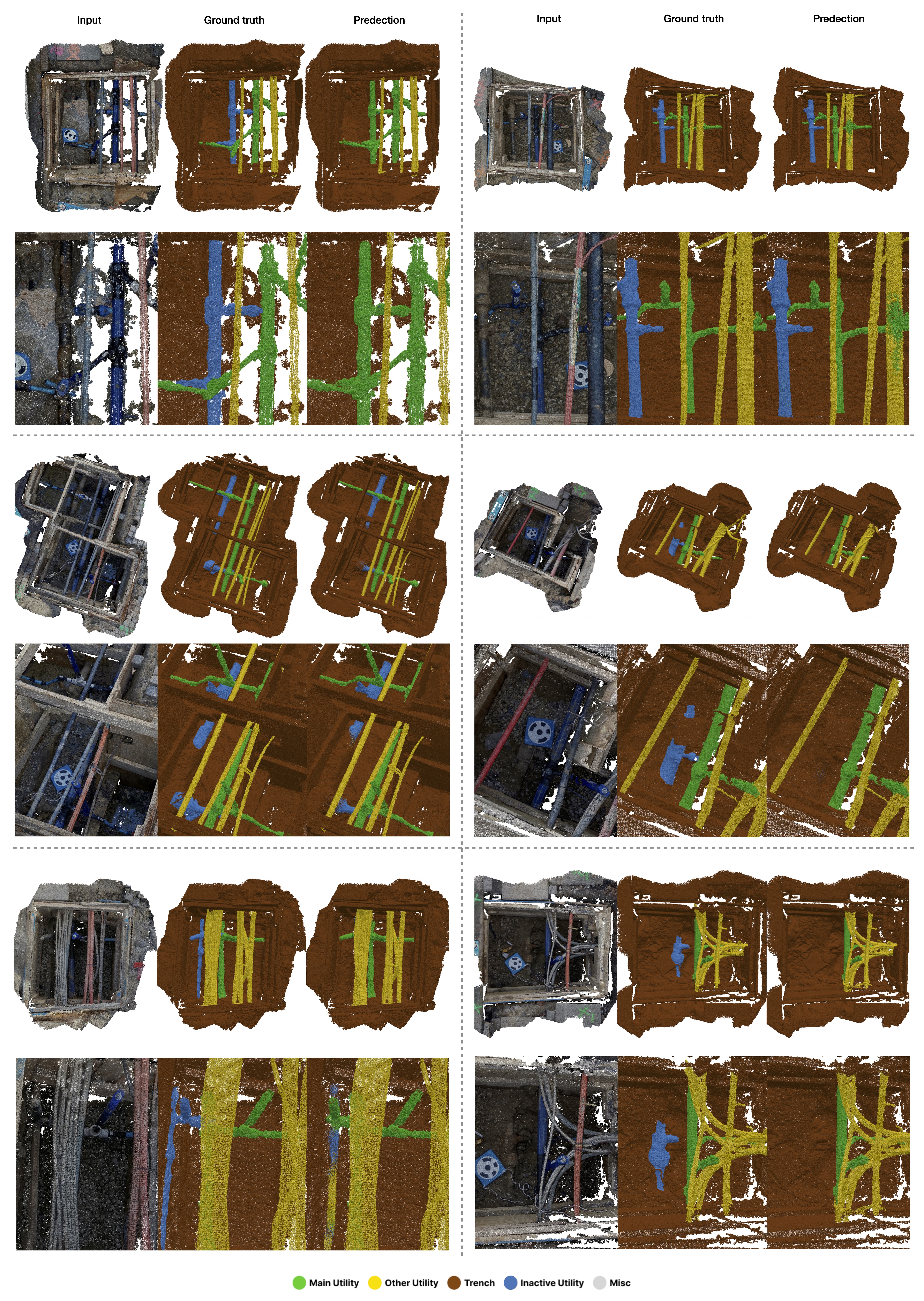}
\caption{Qualitative Examples from running inference on samples from Water Area 2, with a trained PointVector model on Water Area 1, 3, 4 and 5. In this model, the \textit{Inactive Utility} class was included to test against the \textit{Main Utility} and \textit{Other Utility} classes.}
\label{fig:5-fold-inactive-large}
\end{figure*}

\section{Qualitative Examples from Fine-tuning evaluation on Heating Areas}
\label{sec:appendix_d}
We provide additional qualitative examples from running inference on various samples from Heating Area 2 using various trained versions of the PointNeXt model in figure \ref{fig:supplementary-fine-tuning-vs-training-qualitative_examples}. 
These are supplements to the qualitative examples in figure 5 of the main paper. 
We showcase inference results from a PointNeXt models trained from scratch on solely 1 and 10 samples from Heat Area 1 (Baseline), a PointNeXt model trained on samples from Water Area 1-4 (pre-trained) and finally PointNeXt models pre-trained on Water Area 1-4 and further fine-tuned on 1 and 10 samples from Heating Area 1, were only weights of the segmentation head are tuned.

\begin{figure*}[ht]
\centering
\includegraphics[width=1\linewidth]{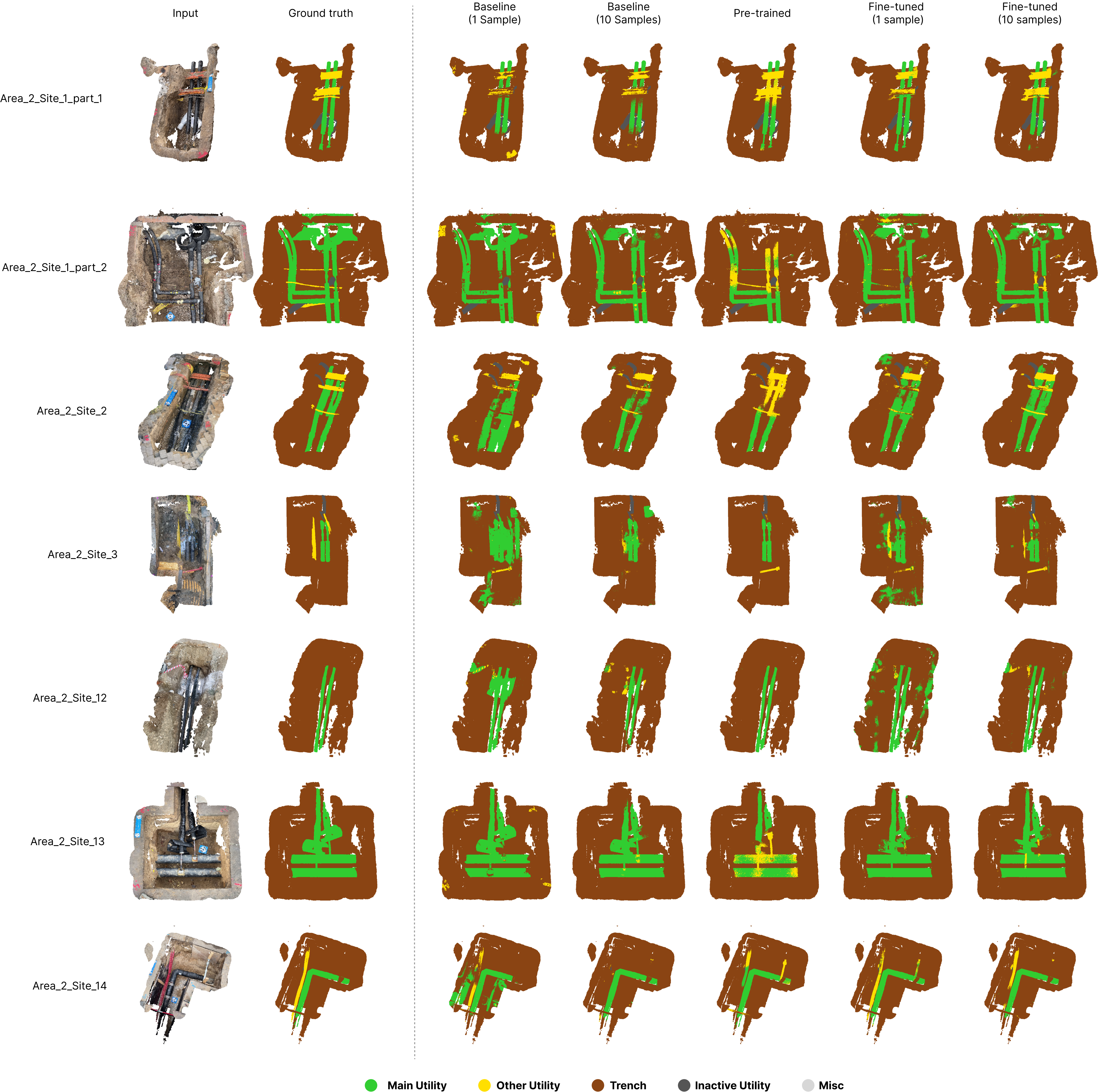}
\caption{Qualitative Examples from running inference on samples from Heating Area 2 with 5 trained PointNeXt model versions: 2 models which are trained on 1 and 10 samples from Heating Area 1 (called Baseline), 1 model pre-trained on samples from Water Area 1-4 and 1 model pre-trained on samples from Water Area 1-4 and fine-tuned with 1 and 10 samples from Heating Area 1.}
\label{fig:supplementary-fine-tuning-vs-training-qualitative_examples}
\end{figure*}

 \fi

\end{document}